\documentclass[conference]{IEEEtran}
\IEEEoverridecommandlockouts
\usepackage{cite}
\usepackage{amsmath,amssymb,amsfonts}
\usepackage{algorithmic}
\usepackage{graphicx}
\usepackage{textcomp}
\usepackage{multirow}
\usepackage{bbding}
\usepackage{hyperref}

\def\BibTeX{{\rm B\kern-.05em{\sc i\kern-.025em b}\kern-.08em
    T\kern-.1667em\lower.7ex\hbox{E}\kern-.125emX}}
\begin{document}

\title{Loss Rank Mining: A General Hard Example Mining Method for Real-time Detectors\\}

\author{\IEEEauthorblockN{Hao Yu$^1$, Zhaoning Zhang$^{1 *}$,\thanks{$^*$Corresponding author. Email: zzningxp@gmail.com} Zheng Qin$^1$, Hao Wu$^2$, Dongsheng Li$^1$, Jun Zhao$^3$, Xicheng Lu$^1$}
\IEEEauthorblockA{$^1$Science and Technology on Parallel and Distributed Laboratory, \\
National University of Defense Technology, Changsha, China\\
$^2$College of Electronic and Engineering, National University of Defense Technology, Changsha, China\\
$^3$College of Meteorology and Oceanology, National University of Defense Technology, Changsha, China\\
}
}
\maketitle

\begin{abstract}

 Modern object detectors usually suffer from low accuracy issues, as foregrounds always drown in tons of backgrounds and become hard examples during training. Compared with those proposal-based ones, real-time detectors are in far more serious trouble since they renounce the use of region-proposing stage which is used to filter a majority of backgrounds for achieving real-time rates. Though foregrounds as hard examples are in urgent need of being mined from tons of backgrounds, a considerable number of state-of-the-art real-time detectors, like YOLO series,  have yet to profit from existing hard example mining methods, as using these methods need detectors fit series of prerequisites. In this paper, we propose a general hard example mining method named Loss Rank Mining (LRM) to fill the gap. LRM is a general method for real-time detectors, as it utilizes the final feature map which exists in all real-time detectors to mine hard examples. By using LRM, some elements representing easy examples in final feature map are filtered and detectors are forced to concentrate on hard examples during training.  Extensive experiments validate the effectiveness of our method. With our method, the improvements of YOLOv2 detector on auto-driving related dataset KITTI and more general dataset PASCAL VOC are over 5\% and 2\% mAP, respectively. In addition, LRM is the first hard example mining strategy which could fit YOLOv2 perfectly and make it better applied in series of real scenarios where both real-time rates and accurate detection are strongly demanded.

\end{abstract}


\section{Introduction}
\label{S:1}

\begin{figure*} 
\centering
\includegraphics[width = 0.95\textwidth]{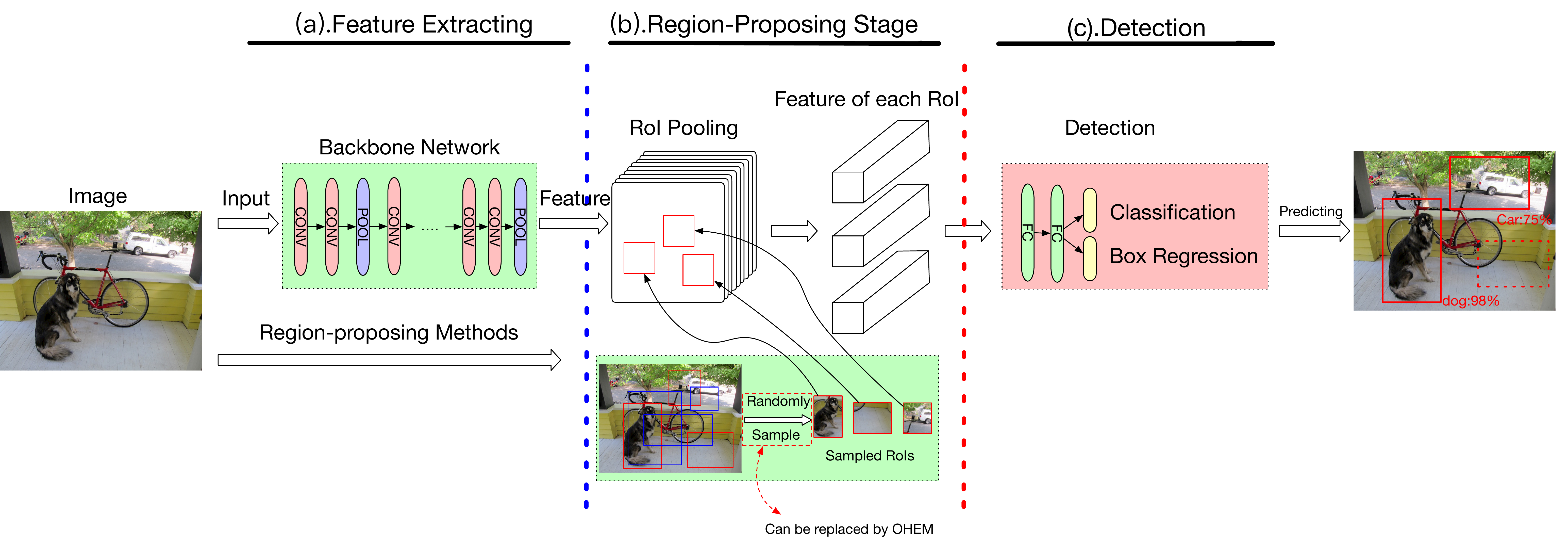}
\caption{The pipeline of a representative detector with region-proposing stage named Fast-RCNN\cite{girshick2015fast} is illustrated in this figure. (a) is the feature extracting stage which uses CNN to extract features for each image; (b) is the region-proposing stage which proposes thousands of RoIs that are likely to contain objects for each image, but only a minority of them are randomly sampled and passed to RoI Pooling layer. Then RoI pooling layer produces feature vectors for each RoI;(c) is the detection stage, which takes RoIs' feature vectors as input and outputs one prediction for each RoI. (a) is a shared network because it computes for the whole image. By contrast, (c) is a RoI-wise network, as it is calculated for each RoI rather than the whole image. Final predictions are made based on the RoIs proposed in (b) and the position of bounding-boxes is adjusted by box regression. In this case, three RoIs are inputed to the detection network, but only two are detected as foregrounds and the rest is considered as a background.}
\label{fig:fig2}
\end{figure*}

The huge imbalance between backgrounds and foregrounds is native to the realm of object detection since millions of regions can be sampled from an image, but only a few are considered as foregrounds. The imbalance issue has exerted tons of side-effects throughout the development of object detectors, as detectors tend to be dominated by backgrounds and fail to detect objects. Previous work\cite{girshick2014rich,girshick2015fast,ren2015faster,dai2016r,he2017mask} adopted the region-proposing stage to mitigate that imbalance.  Region-proposing stage proposes Regions-of-Interest(RoIs) which are more likely to contain objects to narrow the sampling spaces of subsequent detection. However, as that stage proposes thousands of RoIs for each image and only a minority of them are foregrounds, backgrounds still overnumber foregrounds by tens or hundreds to one. To tackle that problem, in some proposal-based detectors\cite{girshick2014rich,girshick2015fast,ren2015faster}, sampling strategies were adopted after RoIs being proposed for further balance.  Take \cite{girshick2015fast} for instance, after RoIs being proposed for an input image, 16 foregrounds and 48 backgrounds are sampled randomly from all RoIs and only these 64 RoIs are used for training. The pipeline of Fast-RCNN detector in \cite{girshick2015fast} is illustrated in Fig.~\ref{fig:fig2}. 

%For further balance, Girshick et al. \cite{girshick2014rich,girshick2015fast} and Ren et al.\cite{ren2015faster} sampled region proposals another time after the region-proposing stage. Afterwards, Shrivastave et.al \cite{shrivastava2016training}  replaced the sampling strategy in \cite{girshick2015fast} with a hard example mining strategy, namely, Online Hard Example Mining(OHEM). Admittedly, those strategies alleviate the side-effects of imbalance significantly and do contribute to the improvements of detection accuracy.  

%Techniques for balancing backgrounds and foregrounds during training has been adopted in early proposal-based detectors\cite{girshick2014rich,girshick2015fast,ren2015faster}. Region-proposing stage proposes Regions-of-Interest(RoIs) which are more likely to contain objects to narrow the sample spaces of subsequent detection. As that stage proposes thousands of RoIs for each image and only a minority of them are foregrounds, backgrounds still overnumber foregrounds by tens even hundreds to one. To tackle that problem, part of foregrounds and negatives are sampled in a fixed ratio. Take \cite{girshick2015fast} for instance, 128 foregrounds and 128 backgrounds are sampled randomly for training. The pipeline of a representative proposal-based detector is illustrated in Fig.~\ref{fig:fig2}. 

Apparently, the sampling strategy mentioned above has a severe shortage. It samples RoIs randomly but in each iteration, different RoIs make diverse contributions to the model and RoIs that contribute most are not always selected by the sampler. To solve this, Shrivastave et al. proposed Online Hard Example Mining(OHEM) to measure contributions of each RoI and sample most beneficial ones for training in each iteration. More specifically, OHEM can be split into three stages. Firstly, all the RoIs proposed by region-proposing stage are used to propagate forward without further sampling, and loss values are calculated for all of them. Then, all the RoIs are ranked in loss-descent order and only a fixed number of RoIs with high loss values are selected. Finally, RoI pooling layer constructs feature vectors for selected RoIs and those feature vectors used for training. OHEM boosts detection accuracy significantly, as it samples RoIs where the model performs worst for training and they tend to provide essential information to the model. Additionally, OHEM as a sampling strategy contributes to mitigating the imbalance between backgrounds and foregrounds significantly since foregrounds as minority incline to possess high loss values and are more likely to be selected by OHEM.

\begin{figure*} 
\centering
\includegraphics[width = 0.95\textwidth]{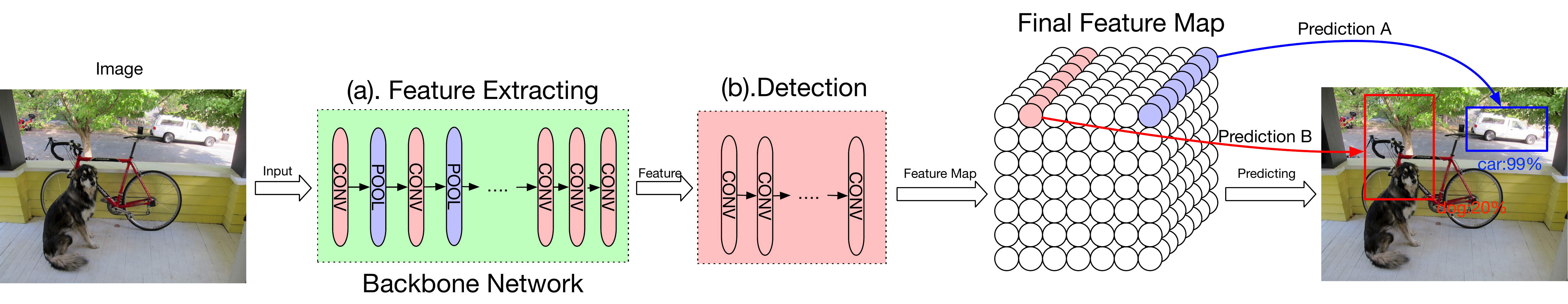}
\caption{The pipeline of a representative real-time detector with region-proposing stage named YOLOv2 is illustrated in this figure. (a) is the backbone network for feature extracting; (b) is the detection network which outputs tons of predictions for each image. It must be noticed that real-time detectors do not adopt a region-proposing stage and do not rely on the presence of RoIs. Real-time detectors use the last feature map to represent tons of predictions that they make. Prediction A and Prediction B are two bounding-boxes predicted by the detector and each of them is presented by a group of output elements in final feature map. In details, output elements in same spatial positions but different channels of the final feature map represents diverse characters of a bounding-box, e.g. coordinates, widths, heights and classes.}
\label{fig:fig5}
\end{figure*}

However, for real-time detectors that regress  straightly to final predictions without using RoIs, OHEM as a sampling strategy itself serves no purpose. The general pipeline of real-time detectors are shown in Fig.~\ref{fig:fig5}. It is clear in Fig.~\ref{fig:fig5} that no RoI is sampled when using real-time detectors to detect objects, as a result of which, RoI-free hard example mining strategies need to be proposed for alleviating the huge imbalance between backgrounds and foregrounds. Liu et al. utilized hard example mining strategies in\cite{liu2016ssd} to select hard negative examples during training and make the ratio of foregrounds to backgrounds 1:3 compulsively. To some extent, this strategy alleviates the imbalance between foregrounds and backgrounds as backgrounds are down-sampled during training. However, all foregrounds are considered as hard examples in \cite{liu2016ssd} and relationships between foregrounds and negatives are totally neglected. To mine foregrounds for real-time detectors more effectively,  Lin et al. proposed Focal Loss\cite{lin2017focal} method. Focal Loss modifies the loss function to increase the gradients produced by hard examples and decrease that calculated for easy ones at the same time. In this way, it boosts the accuracy of real-time detectors significantly. However, Focal Loss is not a general method for real-time detectors as it heavily depends on the definition of loss function.  For instance, Focal Loss can be used to boost the detection accuracy of SSD\cite{liu2016ssd} but it cannot be applied to YOLOv2\cite{redmon2016yolo9000} which adopts a totally different loss function. Accordingly, the fact is that for YOLOv2 detector which is broadly used for real-time detection in many real scenarios, no existing hard example mining strategy could fit it well and help to boost its detection accuracy.%Taking the loss function used in \cite{lin2017focal} for instance, the loss function can be devided into the addition of two sub-functions, where one is the loss of classification and the other is the box regression loss. Focal loss can be applied in the loss of classification perfectly as background serves as a unique class of foreground and the classification loss is calculated for both of them. But for box regression loss, it is calculated for positives only while that of negatives is setted to 0 compulsively. As a result, when trying to alleviate the side-effects of imbalance between foreground and background of YOLOv2, focal loss serves no useful purpose, because YOLOv2 uses more complex multi-task loss with 4 sub-functions, say, loss for object(obj), non-object(non-obj), classificatio,n(cls) and box regression(reg), among which no sub-function is calculated for both foreground and background. More specifically, the loss for foreground is obj + cls + reg while that for background is non-obj only, which means that no common part can be used by focal loss.

%Similar ideas have occurred in previous work. Shrivastava et al. proposed a method to mine foreground from tons of background based on the idea of hard example mining, namely, Online Hard Example Mining(OHEM)\cite{shrivastava2016training}. Using OHEM is independent to the definition of loss functions, as it does not modify the loss function but utilizing final loss values to select hard examples. Details of OHEM are given in Fig.\ref{fig:fig2} and the Fast-RCNN detector in \cite{girshick2015fast} serves as the basement of OHEM. It is shown clearly in Fig.\ref{fig:fig2} that OHEM relies on Regions-of-Interest(RoIs) proposed by region-proposing stage which was broadly adopted by previous work\cite{girshick2014rich,girshick2015fast,ren2015faster,dai2016r,he2017mask} but removed in state-of-the-art real-time detectors\cite{liu2016ssd,redmon2016you,redmon2016yolo9000} for higher speed. OHEM acts as a sampler which selects a fixed number of RoIs in each iteration according the final loss values and only these RoIs are used to train the model. Accordingly, OHEM performs on RoIs and depends on the existence of region-proposing stage . For real-time detectors whose region-proposing stage is removed, OHEM serves no purpose. 

In this paper, to tackle existing problems that we have mentioned above, we propose the \textbf{L}oss \textbf{R}ank \textbf{M}ining(LRM) method to mine foregrounds effectively for real-time detectors. To the best of our knowledge, LRM method is the first general hard example mining method which could be applied to all state-of-the-art real-time detectors. For YOLOv2 detector which performs well on many real scenarios but have yet to be optimized by existing hard example strategies, our LRM method could help to boost its detection accuracy. Compared with \cite{lin2017focal}, we modify the final feature map which is a general structure used in all real-time detectors to represent predictions rather than adjusting loss functions which can be defined diversely in various real-time detectors. The principle of LRM is that output elements representing predictions with low loss values are filtered before backpropagation, and only predictions with high loss values contribute to training the detectors. To sum up, the contributions of our LRM method can be concluded as follows:

\begin{itemize}
\item LRM is the first general hard example mining method which could be applied to all state-of-the-art real-time detectors for higher detection accuracy, without any side-effects on real-time detection rates.
%\item LF utilizes the thought of Dropout to mine hard examples. It mines foregrounds from backgrounds and prevents model from overfitting in one go. 
\item LRM successfully boosts detection accuracy of YOLOv2 detector which has been broadly adopted in many real-time detection scenarios but yet to be optimized by existing hard example mining methods.

\end{itemize}

The subsequent parts of this paper are organized as follows. Some previous efforts are introduced in Section~\ref{S:2}. Then, our approach is detailed in Section~\ref{S:3}. Next, experiment results show the effectiveness of our approach in Section~\ref{S:4}. Finally, Section~\ref{S:5} concludes our work.

\section{Related Work}
\label{S:2}

\begin{figure*} 
\centering
\includegraphics[width = .8\textwidth]{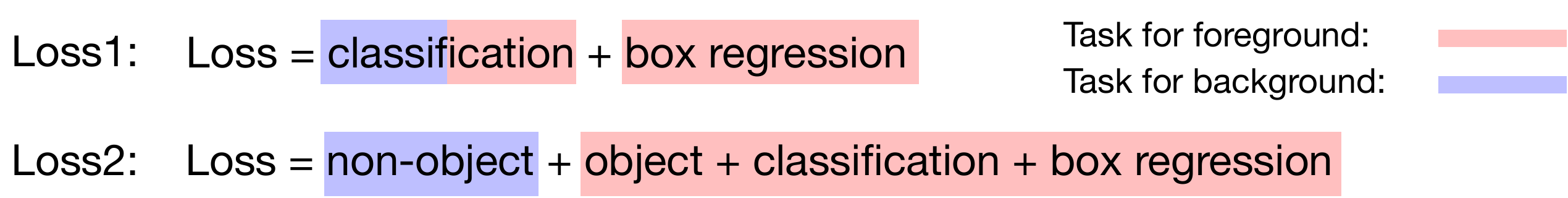}
\caption{Two widely-used muti-task loss functions in object detection are illustrated in this figure. Previous work like\cite{girshick2014rich,girshick2015fast,ren2015faster} adopt Loss1 and \cite{lin2017focal} is also based on it. For methods like \cite{redmon2016you,redmon2016yolo9000}, they use Loss2 with 4 tasks, namely, loss for object(obj), non-object(non-obj), classification(cls) and box regression(reg)}
\label{fig:fig1}
\end{figure*}
\subsection{Traditional Hard Example Mining Techniques}
\label{S:2:s:1}

Many kinds of hard example mining techniques have been widely applied to training classic models\cite{suykens1999least,dollar2009integral}. 

Boosted decision tree in \cite{dollar2009integral} is trained with hard example mining strategy but hard examples are mined one time only. To begin with, all the positive examples and a random set of negative examples are blended together as the original training set. After reaching convergence on the original training set, the trained model is applied to the rest of negative examples. Then, only false positive examples are selected as hard ones and added to the original training set to form the final training set. Finally, the model is trained on the refreshed training set until convergence. 

Another hard exampling mining technique named boostrapping is used to train Support Vector Machines(SVMs)\cite{suykens1999least} and hard examples are mined several times in this case. A working set containing a tiny number of samples is used in boostrapping. Samples are added to and removed from this set according to some specific rules. Processes of training model to convergence on the existing working set and utilizing the trained model to modify the working set are finished alternatively. When modifying the working set, samples in the working set classified correctly by the existing model are removed from it while samples out of the working set misclassified by the model are added to it.

\subsection{Hard Example Mining in Early Object Detectors}
\label{S:2:s:2}

%Previous work has noticed that object detection methods are faced with a huge gap between the number of positives and that of negatives, which has became the bottleneck preventing detectors from further advancement. As a result, a variety of hard exampling mining techniques have been adopt in the realm of object detection.

The history of utilizing techniques of hard example mining in object detection can date back to the time when it was used to train SVMs for pedestrian detection\cite{dalal2005histograms}. After the prevalence of CNN-based model in object detection, hard example mining still played an important role as an SVM classifier is usually attached to the top of detectors for classification, e.g.\cite{girshick2014rich, he2014spatial}. 

However, after SVMs being replaced by layers consisting of neural units in subsequent object detection methods\cite{girshick2015fast,ren2015faster}, hard example mining strategies were not utilized in the training of CNN-based detectors until Online Hard Example Mining(OHEM) proposed in \cite{shrivastava2016training}. OHEM depends on the RoIs proposed by the region-proposing stage heavily but that stage is removed in state-of-the-art real-time detectors for higher speed, which makes OHEM serve no purpose on those detectors. 

\subsection{Focal Loss for Real-time Detectors}
\label{S:2:s:4}

To tackle severe imbalance issues between backgrounds and foregrounds in real-time detectors, Lin et al. proposed Focal Loss\cite{lin2017focal} and tried to modify the loss function to mine the hard examples from easy ones. However, Focal Loss depends on the definition of loss function heavily and cannot be applied to plenty of state-of-the-art real-time detectors straightly. 

More specifically, Focal Loss needs the whole loss function or a specific part of it to be calculated for both foregrounds and backgrounds. The shared part of the loss function is multiplied with a self-adaptive parameter to emphasize foregrounds and de-emphasize backgrounds. In cases where the unshared part dominates the whole loss function or both sides share no part, Focal Loss is less effective or even serves no purpose. 

Two multi-task loss functions broadly adopted in object detection are illustrated in Fig.\ref{fig:fig1}. Loss1 consists of two tasks, namely, classification loss and box regression loss. Classification loss in Loss1 is calculated for both foregrounds and backgrounds, but box regression loss is computed for foregrounds only. Though classification loss is the majority of Loss1 and Focal Loss can be applied to it, the impacts produced by box regression loss are totally neglected. Additionally, for the methods in \cite{redmon2016you,redmon2016yolo9000} which adopt Loss2 as their loss functions, Focal Loss serves no purpose. This is mainly because Loss2 possesses four subtasks, namely, object loss, non-object loss, classification loss and box regression loss, but all of them cannot be shared by both foregrounds and backgrounds. Though replacing Loss2 with Loss1 makes Focal Loss available for those detectors, it does harm to detection accuracy significantly, as Loss2 fit those detectors better. Our experiments in Section \ref{S:4:s:5} demonstrate that for those detectors, applying Focal Loss compulsively is detrimental but using Loss Rank Mining is helpful.

\begin{figure*} 
\centering
\includegraphics[width = 0.9\textwidth]{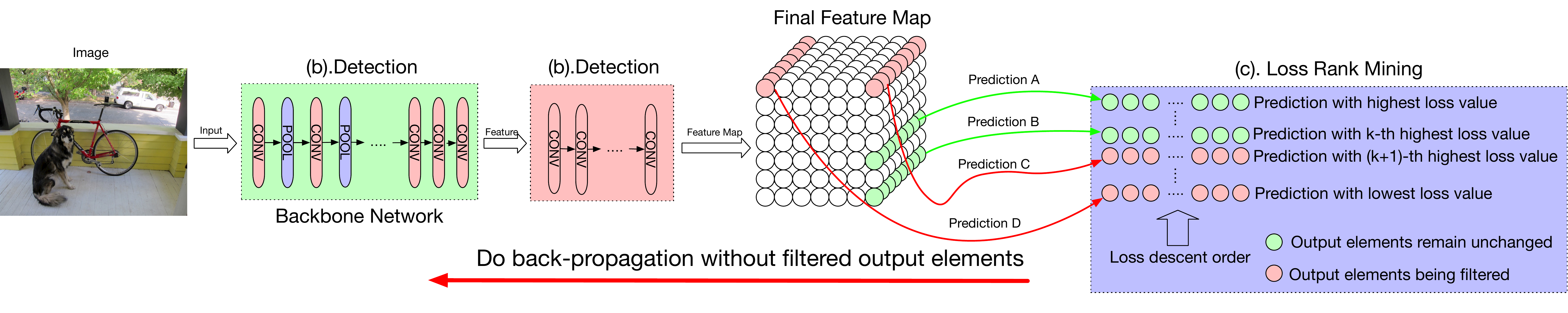}
\caption{The pipeline of training a  real-time detector with LRM module is shown in this figure. (a) and (b) are as same as those in Fig.~\ref{fig:fig5}. (c) illustrates the LRM module on real-time detectors. Output elements are reorganized and grouped according to regions they represented for hard example mining. Prediction A, B, C and D each presented by a group of output elements represent 4 bounding-boxes predicted by the detector. In LRM module, total N predictions are ranked in loss-descent order and K regions with highest loss are selected as hard examples for backpropagation. Take these 4 predictions for instance, Prediction A and B are remained but the other two predictions are filtered away. Output elements belonging to Prediction A and B remain open while that representing Prediction C and D are closed and prevented from backpropagation.}
\label{fig:fig3}
\end{figure*}

% \usepackage{multirow}

\section{Approach}
\label{S:3}
In this section, we aim at detailing the LRM method that we propose to improve detection accuracy of real-time detectors. To begin with, Section~\ref{S:3:s:1} details the pipeline of our LRM method. Then, we introduce how to use LRM in training and inference in Section~\ref{S:3:s:2}.

\subsection{Loss Rank Mining  Method for Real-time Detectors}
\label{S:3:s:1}

%For YOLOv2, the strategy used in Fast-RCNN is not suitable, but fortunately, we have another simple but efficient method.

%According to the first 2 dominating preconditions mentioned in Section~\ref{S:3:s:1}, strategies used in \cite{shrivastava2016training} is not suitable for YOLOv2. On the one hand, there are no RoIs proposed in YOLOv2 and predictions are made by straight regression. On the other hand, YOLOv2 is a blended structure for the beginning to the end and computation is shared by all of the final predictions. So we must turn to another way to implement OHEM on YOLOv2. The method is setting the gradient of easy examples to 0(called zero-gradient strategy), which is abandoned by \cite{shrivastava2016training} but adopted by our work. 

For constructing a general hard example mining method for real-time detectors, common grounds shared by all the existing state-of-the-art real-time detectors must be utilized. More specifically, common points of real-time detectors can be concluded into 2 aspects. Firstly, all of them abandon the use of region-proposing stage. The computation of all potential region proposals is blended as well as implicit. No region proposal is split from whole images before final predictions being made.  Secondly, for one input image, tons of predictions are made by real-time detectors in one go, and final feature map(or a group of final feature maps\cite{lin2016feature}) is used to represent all the predicted bounding-boxes. According to the aforementioned two points, a general hard example mining method for real-time detectors must be a RoI-free one and the final feature map as a general structure should be used to filter easy examples. Accordingly, our strategy is filtering some output elements which represent well-detected bounding-boxes in final feature map for the purpose of concentrating on hard examples.

The pipeline of Loss Rank Mining method can be concluded as follows： (a). For an input image, use a backbone network to extract its representative features and forward them to detection phase, and then, obtain final feature map which represents all the final predictions; (b). Reorganize and group output elements in the final feature map according to which predictions that they represent; (c). Calculate loss values for each prediction, and rank them in a loss-descent order;(d). Select top-K predictions with highest loss values and filter all the output elements that represent outer predictions. (e). Do backpropagation without gradients of those output elements which has been filtered. Details are shown in Fig.~\ref{fig:fig3}. Moreover, non-maximum suppression(NMS)\cite{girshick2015fast} is used after all predictions being ranked, as co-located predictions with high Intersection-over-Union(IoU) serve similar functions during backpropagation and selecting them as hard examples for multiple times is meaningless.

%More specifically, reasons why we do this are mainly consisted of 3 aspects, the first is that there are no RoIs proposed on YOLOv2 which means we do not known the possible positions of objects until making final predictions, so OHEM must perform on the final predictions instead of RoIs; the second is that YOLOv2 is a computation-blended network from the very beginning to the end, which means though considered as easy examples, specific final predictions cannot be removed in any phase of this network; the last but not least is that though gradients are filled with tons of 0 in the very beginning of back-propagation, the computation does not increase significantly. The computation in forward phase increases slightly due to the ranking of all the predictions, but that in backward phase remains unchanged. Accordingly, using zero-gradient strategy is a suitable and effective way to implement OHEM on YOLOv2. 
\begin{table}
\centering
\caption{Flops and mAP comparisons between different backbone networks}
\label{tab:tab2}
\begin{tabular}{|l|l|l|l|}
\hline
Detector                & Backbone     & Flops    & mAP  \\ \hline
\multirow{3}{*}{YOLOv2} & Darknet      & 34.90 Bn & 76.8 \\ \cline{2-4} 
                        & Tiny Darknet & 6.97 Bn  & 57.1 \\ \cline{2-4} 
                        & MobileNet    & 2.56 Bn  & 67.5 \\ \hline
\end{tabular}
\end{table}
\subsection{Training and Inference with LRM Method}
\label{S:3:s:2}

Like previous hard example mining strategies, Loss Rank Mining method is also a technique used in training phase only. In training phase, we add a LRM module after the final feature map. In LRM module, detectors' final feature map \textbf{F} are multiplied by a mask matrix \textbf{M} of same size in an elementwise way. The mask matrix is a binary one whose values depends on whether the corresponding elements in \textbf{F} belong to a hard example. The operation in LRM module can be illustrated by the following equation:
\begin{equation} 
\label{eq:eq1}
\hat{F} = F \circ  M,
\end{equation}
where \textbf{$\hat{F}$} is the output of LRM module and \textbf{$\circ$} is the symbol of elementwise product.

The values of mask matrix \textbf{M} are not fixed but they do not need to be learned by the model. As we have mentioned in Section~\ref{S:3:s:1}, these values are decided by final predictions. At the beginning of each iteration, all the elements in \textbf{M} are set to 0 and after \textbf{F} being calculated, for output elements belonging to hard examples, the corresponding elements in \textbf{M} are set to 1. Then calculation of \textbf{$\hat{F}$} is finished for backpropagation. By using $\hat{Loss}$ for backpropagation, the partial derivative of $\hat{Loss}$ with respect to any element in \textbf{F} can be calculated as follows:

\begin{equation}  
\label{eq:eq2}
\frac{\partial{\hat{Loss}}}{\partial{f_{i,j}^{c}}}=m_{i,j}^{c} \cdot \frac{\partial{\hat{Loss}}}{\partial{\hat{f_{i,j}^{c}}}}=
\left\{  
             \begin{array}{lr}  
             0, &  m_{i,j}^{c} = 0,\\  
             \frac{\partial{Loss}}{\partial{f_{i,j}^{c}}} & m_{i,j}^{c} = 1,
             \end{array}  
\right.
\end{equation}

where $f_{i,j}^{c}$ represents an element in  \textbf{F} whose position is (i, j) and channel is c. $m_{i,j}^{c}$ is the corresponding element in \textbf{M}. It is shown clearly that by adopting the mask matrix \textbf{M}, output elements belonging to easy examples are prevented from backpropagation successfully.

 LRM method is only used in the training phase to make the model concentrate on hard examples and adjust the direction of gradient descent to reach a better solution. For utilizing the trained model to detect objects, the LRM module is removed and the original architecture is adopted. Accordingly, adopting LRM method does not introduce any additional time consumption in the inference phase, which could boost the detection accuracy while maintaining the real-time rates.

\begin{figure} 
\centering
\includegraphics[width = .4\textwidth]{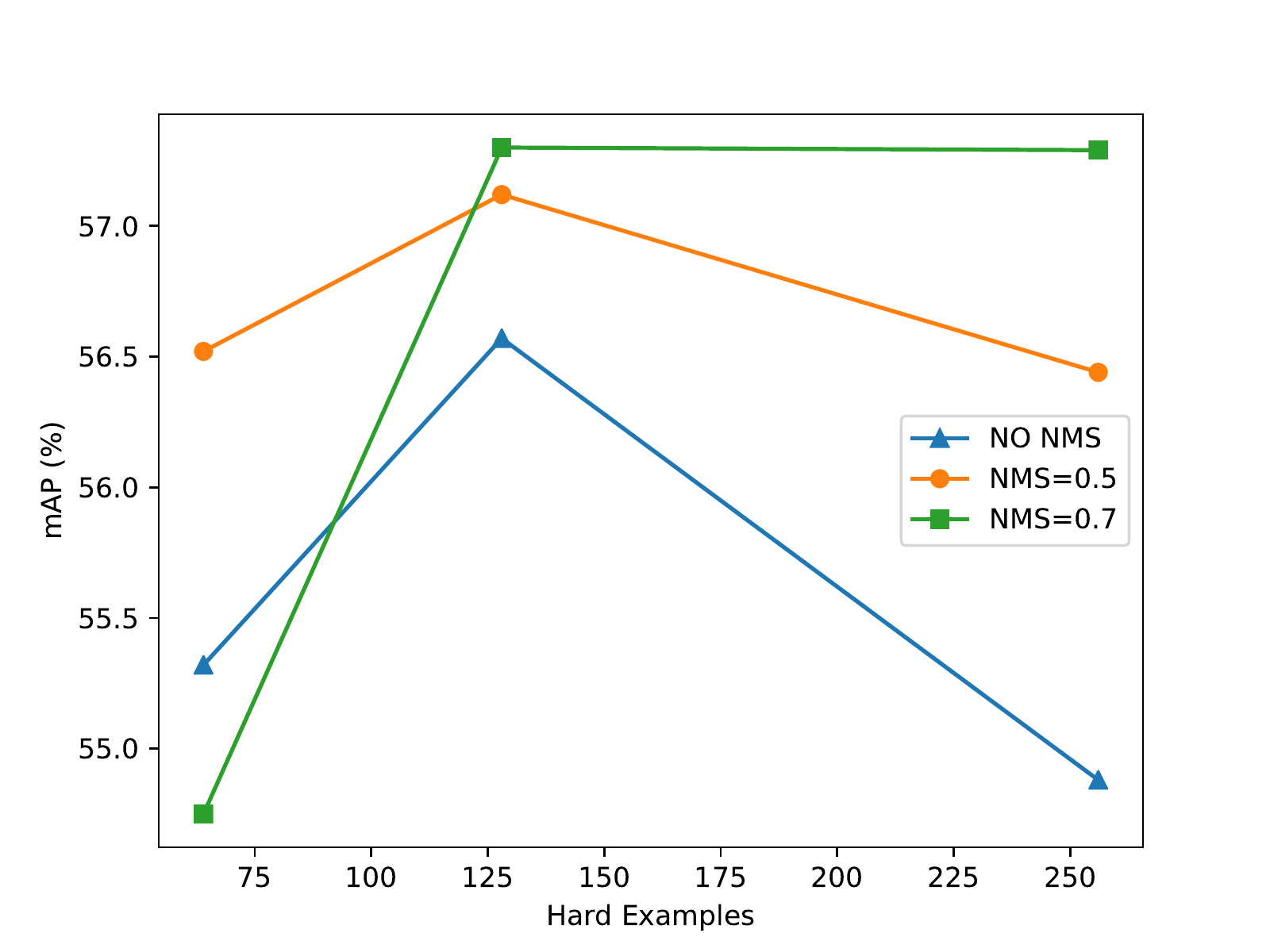}
\caption{mAP of different NMS thresholds when changing the number of hard examples on KITTI dataset.}
\label{fig:fig4}
\end{figure}

\begin{figure} 
\centering
\includegraphics[width = .4\textwidth]{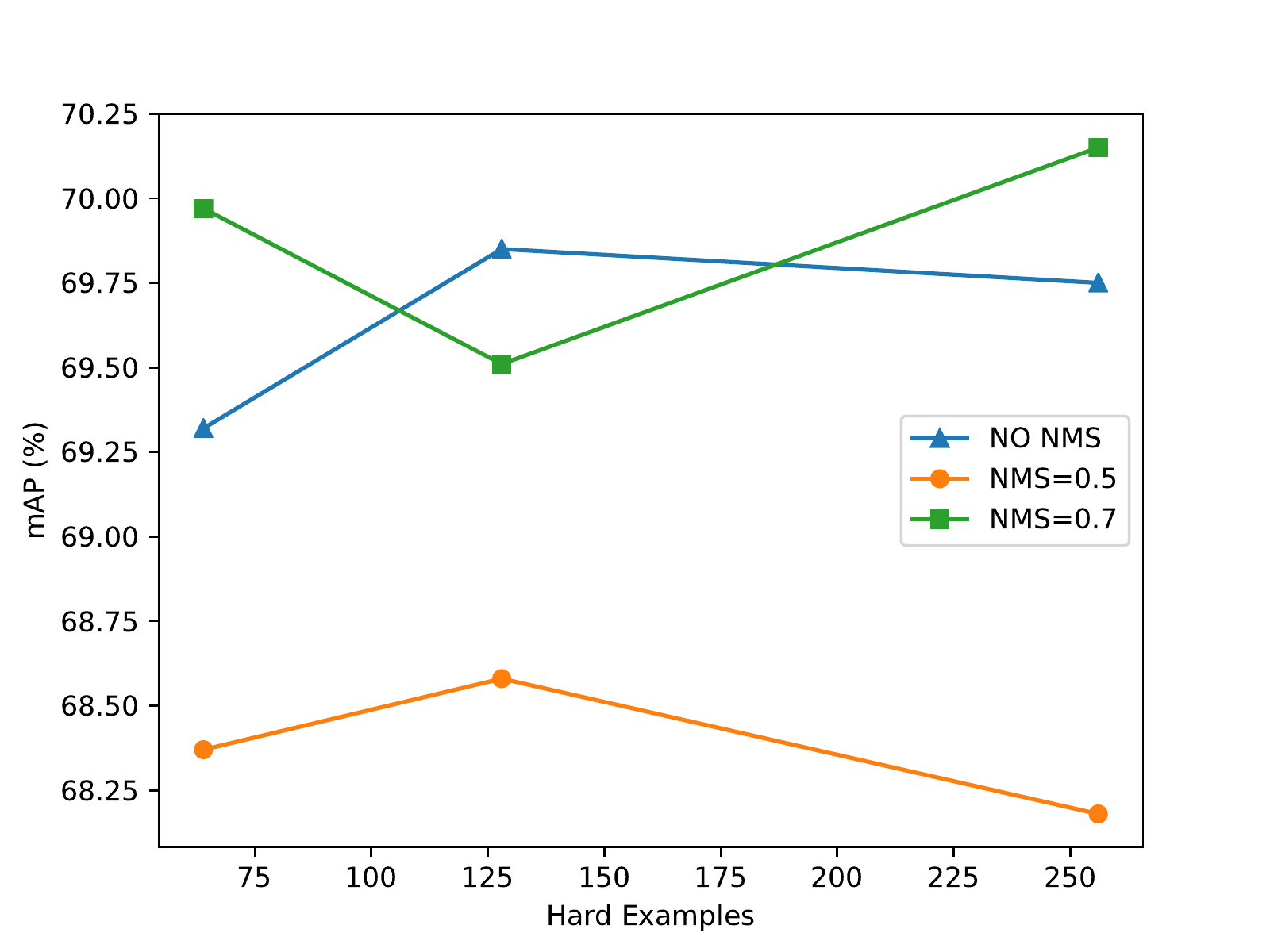}
\caption{mAP of different NMS thresholds when changing the number of hard examples on PASCAL VOC dataset.}
\label{fig:fig6}
\end{figure}

\section{Experiment}
\label{S:4}{}
\subsection{Experimental Setup}
\label{S:4:s:1}

All the experiments are implemented on lightweight deep learning framework Darknet\cite{redmon2013darknet} and run on several K40 GPUs. We only utilize single GPU to train a single model rather than adopting muti-GPU training strategies.

Two datasets are utilized to validate the effectiveness of our model, namely, KITTI\cite{geiger2013vision} and PASCAL VOC\cite{everingham2010pascal}. KITTI dataset consists of tons of objects related to the driving scenario, such as pedestrians, cars, cyclists and so forth. We set experiments on it to show that some real-time applications, like auto-driving, can profit a lot from our LRM method. Compared with KITTI, PASCAL VOC is a larger datasets and it includes objects belonging to 20 classes. It is a more general dataset and it was used to prove that LRM can improve the accuracy of real-time detectors in some more complicated scenarios. 

In details, 6 classes in KITTI dataset are used in our experiments, say, car, truck, tram, van, pedestrian and cyclist. Training set and testing set are separated randomly by a ratio of 9:1. For training on PASVAL VOC, the combination of VOC2007 trainval and VOC2012 trainval are used as the training set, and VOC2007 test is used to evaluate the performance of our LRM method. All the experiments use models pre-trained on 1K-ImageNet\cite{deng2009imagenet} and detectors are trained 45K iterations on PASCAL VOC. The resolution of input images is set to 416*416 and batch size is set to 64.

We use 2 hyperparameters to adjust our models and 10 models are constructed on each dataset, including the original detector which serves as the baseline. The first hyperparameter is the number of hard examples. This hyperparameter represents the number of predictions remained for backpropagation. The other hyperparameter is the NMS threshold, which is the limitation we put on the final predictions. The NMS method is used to remove some redundant informations. More specifically, when the IoU(Intersection-over-Union) of two predictions belonging to same classes is equal or lager than the threshold, only the one with higher loss values should be remained. When use smaller NMS threshold, stricter limitation is introduced and more redundant informations are removed.
\begin{table*}[!htbp]
\centering  % 表居中
\setlength{\tabcolsep}{4pt}
\caption{Results on KITTI dataset.}
\label{tab:tab3}
%\resizebox{0.6\textwidth}{0.7in}{
\begin{tabular}{|c|c|c|c|c|*{7}{c}|}  % {lccc} 表示各列元素对齐方式，left-l,right-r,center-c
\hline

Detector &Backbone &LRM &Hard examples &NMS &mAP &bike &van	&tram &car &person &truck\\ \hline  

YOLOv2&MobileNet&\XSolidBrush &- &- &51.98	&32.10	&49.56	&67.55	&56.65	&31.61	&74.39\\ 
\hline
\multirow{3}*{YOLOv2}
&\multirow{3}*{MobileNet}
&\Checkmark &64 &\XSolidBrush &55.32	&38.10	&56.50	&67.93	&\textbf{60.87}	&34.96	&73.55\\
& &\Checkmark	&128 &\XSolidBrush	&56.57	&41.57	&56.40	&73.26	&59.55	&34.85	&73.77\\
& &\Checkmark	&256	&\XSolidBrush	&54.88	&39.29	&51.42	&71.09	&59.96	&35.51	&72.00\\
\hline
\multirow{3}*{YOLOv2}
&\multirow{3}*{MobileNet}
&\Checkmark &64	&0.5	&56.52 &39.53 &56.76 &78.08 &58.53 &35.30 &70.92\\
& &\Checkmark	&128 &0.5	&57.12	&36.99	&53.79	&\textbf{82.16}	&59.05	&35.39	&\textbf{75.32}\\
& &\Checkmark	&256	&0.5	&56.44	&38.59	&56.26	&78.00	&59.37	&34.46	&71.98\\
\hline
\multirow{3}*{YOLOv2}
&\multirow{3}*{MobileNet}
&\Checkmark	&64 &0.7	&54.75	&39.80	&55.27	&68.55	&58.70	&\textbf{37.95}	&68.22\\
& &\Checkmark	&128	&0.7	&\textbf{57.30}	&\textbf{41.62}	&56.71	&77.00	&60.06	&34.91	&73.49\\
& &\Checkmark	&256	&0.7	&57.29	&36.85	&\textbf{57.38}	&79.45	&59.84	&36.41	&73.82\\
\hline

\end{tabular}
%}
\end{table*}
\subsection{Model Selection}
\label{S:4:s:2}
In our experiments, we adopt the YOLOv2 detector and apply LRM to it to obtain higher accuracy. Reasons why we choose this detector are various. Firstly, YOLOv2 is an excellent and prevalent real-time object detector which achieves mAP of 76.8\% on PASCAL VOC with FPS of 67.\cite{redmon2016yolo9000} Moreover, like other state-of-the-art real-time detectors, YOLOv2 faces the huge imbalance between backgrounds and foregrounds, but the aforementioned hard example mining methods including  OHEM and Focal Loss serve no purpose on it as YOLOv2 removes region-proposing stage and adopt Loss2 in Fig.~\ref{fig:fig1} as its loss function.  Thus, using LRM to improve the accuracy of YOLOv2 detector is of great significance and all the experiments are based on a YOLOv2 detector.

For using YOLOv2 detector, a suitable backbone network must be chosen. The original version of YOLOv2\cite{redmon2016yolo9000} adopt Darknet as it backbone network, and another implementation of YOLOv2 use a shallower Darnet named Tiny Darknet as its backbone network to decrease the amount of computation and obtain higher detection rates. Moreover, in the recent literature, a considerable number of efforts have been made in building compact neural networks to obtain higher inference speed e.g.\cite{wang2016factorized, iandola2016squeezenet, wu2016quantized, howard2017mobilenets, zhang2017shufflenet}. As these compact networks have achieved great success in the field of image classification, we also adopt a representative one named MobileNet\cite{howard2017mobilenets} as one of the candidate backbone networks. We use these aforementioned backbone networks to construct three YOLOv2 detectors and corresponding comparisons are illustrated in TABLE~\ref{tab:tab2}. In TABLE~\ref{tab:tab2}, the second column is Flops which  represents the amount of computation of detectors and the third column is mAP which represents the detection accuracy on PASCAL VOC dataset. 

It is clearly illustrated in TABLE~\ref{tab:tab2} that MobileNet-based YOLOv2 detector uses least amount of computation(2.56 billion Flops) but still achieves competitive mAP(67.5\%). It is no doubt that compared with the other two backbone networks, MobileNet is a more suitable one for constructing accurate and fast YOLOv2 detector. Thus, a MobileNet-based YOLOv2 detector is used as the baseline in our experiments.

\subsection{Results on KITTI}
\label{S:4:s:3}

This part of experiments is used to validate the effectiveness of LRM method on auto-driving scenario. We set series of experiments on auto-driving related KITTI dataset to prove that LRM method could optimize MobileNet-based YOLOv2 detector and make it more accurate to detecting auto-driving related objects, such as persons, cars, bikes and so forth. The results are illustrated in TABLE~\ref{tab:tab3}.

In this part of experiments, we test 9 MobileNet-based YOLOv2 detectors. All of them are optimized by LRM method but with different hyperparameters. Besides, a MobileNet-based YOLOv2 detector without any additional optimization is used as the baseline. Totally, we construct 10 diverse MobileNet-based YOLOv2 detectors to test the performance of our LRM method. When setting the number of hard examples to 128 and the NMS threshold to 0.7, LRM method achieves the highest mAP on KITTI dataset, which is 57.30\%. Compared with the baseline whose mAP is 51.98\%, LRM method gains a significant improvement of over 5\% mAP. 

More specifically, among all classes, the most significant improvements is achieved on class tram(82.16\% vs. 67.55\%, over 14\% improvements compared with baseline). To achieve this, the number of hard examples is set to 128 and NMS threshold is set to 0.5. But referring to the class truck, the improvement is not so optimistic, even some models' AP on class truck drops moderately. So, we can say that for different classes of data, the effectiveness of our LRM method is diverse. 

We draw Fig.~\ref{fig:fig4} to illustrate the impacts brought by different hyperparameters more clearly. Referring to the number of hard examples, our LRM method achieves the highest mAP when setting the number of hard examples to 128 regardless of what NMS threshold we adopt. When we set the number of hard examples to 64, setting NMS threshold to 0.5 is the best choice. This is mainly because when we only remain a small part of predictions, we should select most beneficial predictions first. However, compared with redundant predictions with higher loss values, these predictions may possesses lower loss values and not be remained. In this case, a smaller NMS threshold should be used to remove those redundant predictions and ensure that most beneficial ones are remained.   But cases become totally different when increasing that number to 128 or 256. Setting NMS threshold to 0.7 achieves the highest mAP in those cases. Because when the number of hard examples grows, we still have many seats after choosing all most beneficial predictions. Some predictions with little redundancy should be remained in this case, as they still can provide some useful informations. Accordingly, a milder NMS threshold should be adopted, for remaining the moderately redundant predictions while removing those heavily redundant ones.

\begin{table*}[!htbp]
\centering  % 表居中
\setlength{\tabcolsep}{2pt}
\caption{Results on PASCAL VOC. VOC2007trainval + VOC2012trainval for training and VOC2007test for testing. }
\label{tab:tab4}
\resizebox{1\textwidth}{0.7in}{
\begin{tabular}{|c|c|c|c|c|*{21}{c}|}  % {lccc} 表示各列元素对齐方式，left-l,right-r,center-c
\hline

Detector &Backbone &LRM &Hard examples &NMS &mAP &aero &bike &bird &boat &bottle &bus &car &cat &chair &cow &table &dog &horse &mbike &person &plant &sheep &sofa &train &tv\\ \hline  

YOLOv2 &MobileNet&\XSolidBrush &- &- &68.00	&70.10	&78.80	&66.67	&53.23	&31.61	&75.79	&74.67	&\textbf{88.90}	&42.61	&71.96	&69.70	&\textbf{85.03}	&86.49	&80.80	&69.68 &34.49	&64.14	&70.00	&80.06	&65.21\\ 
\hline
\multirow{3}*{YOLOv2}
&\multirow{3}*{MobileNet}
&\Checkmark &64 &\XSolidBrush &69.32	&70.60	&80.44	&66.78	&53.13	&34.59	&80.13	&76.64	&86.66	&46.25	&\textbf{74.96}	&71.77	&80.83	&86.35	&81.94	&72.25	&36.20	&66.21	&74.44	&80.34	&65.79\\ 
& &\Checkmark	&128 &\XSolidBrush	&69.85	&71.01	&80.53	&66.28	&\textbf{57.84}	&\textbf{36.27}	&80.42	&76.58	&87.54	&47.04	&71.78	&\textbf{72.80}	&83.34	&84.63	&81.13	&\textbf{72.57}	&34.16	&65.69	&76.13	&\textbf{83.17}	&68.06\\ 
& &\Checkmark	&256	&\XSolidBrush	&69.75	&70.32	&80.65	&66.79	&55.18	&35.15	&79.99	&76.37	&86.32	&46.88	&73.27	&72.22	&82.43	&86.35	&81.37	&71.95	&36.84	&66.27	&75.21	&81.83	&69.71\\
\hline
\multirow{3}*{YOLOv2}
&\multirow{3}*{MobileNet}
&\Checkmark &64	&0.5	&68.37	&69.13	&80.32	&64.42	&54.08	&32.40	&79.09	&76.55	&85.10	&46.82	&70.37	&70.23	&80.19	&84.92	&79.62	&72.03	&32.90	&\textbf{67.40}	&74.44	&81.39	&66.02\\
& &\Checkmark	&128 &0.5	&68.58	&71.51	&81.35	&65.59	&52.35	&35.35	&76.78	&76.46	&84.66	&45.13	&71.54	&70.55	&81.52	&85.39	&80.53	&71.49	&35.59	&65.86	&73.17	&79.88	&66.95\\
& &\Checkmark	&256	&0.5	&68.18	&68.75	&77.14	&62.3	&54.72	&32.97	&77.37	&75.67	&84.52	&45.53	&73.86	&69.98	&81.37	&85.16	&79.75	&71.53	&36.11	&64.44	&75.99	&81.74	&64.79\\
\hline
\multirow{3}*{YOLOv2}
&\multirow{3}*{MobileNet}
&\Checkmark	&64 &0.7	&69.97	&70.20	&81.64	&65.60	&55.71	&35.17	&\textbf{81.29}	&77.62	&86.48	&45.74	&74.83	&70.83	&81.64	&\textbf{87.17}	&\textbf{83.01}	&72.41	&37.57	&67.24	&73.53	&81.68	&\textbf{70.15}\\
& &\Checkmark	&128	&0.7	&69.51	&71.48	&80.84	&65.03	&56.74	&34.76	&79.19	&\textbf{78.01}	&88.34	&47.10	&70.59	&72.47	&82.03	&85.91	&80.90	&71.60	&\textbf{38.85}	&63.20	&71.10	&82.04	&69.98\\
& &\Checkmark	&256	&0.7	&\textbf{70.15}	&\textbf{72.37}	&\textbf{83.27}	&\textbf{66.82}	&55.31	&34.29	&80.92	&76.88	&87.59	&\textbf{48.39}	&74.50	&71.86	&81.80	&86.67	&80.32	&72.45	&37.22	&65.32	&\textbf{76.70}	&82.81	&67.55\\
\hline

\end{tabular}
}
\end{table*}

\begin{table*}[!htbp]
\centering
\caption{Comparisons between Loss Rank Mining and Focal Loss on Darknet-based YOLOv2 detector}
\label{tab:tab5}
\begin{tabular}{|l|l|l|l|l|l|l|l|}
\hline
Detector                & Backbone                 & Loss Function & FL Method & LRM Method & Dataset                     & mAP   & Improvements \\ \hline
\multirow{4}{*}{YOLOv2} & \multirow{4}{*}{Darknet} & Loss2 in Fig.~\ref{fig:fig1}         & \XSolidBrush         & \XSolidBrush                   & \multirow{4}{*}{PASCAL VOC} & 75.13 & -            \\ \cline{3-5} \cline{7-8} 
                        &                          & Loss1 in Fig.~\ref{fig:fig1}     & \XSolidBrush         & \XSolidBrush                   &                             & 73.05 & -2.08        \\ \cline{3-5} \cline{7-8} 
                        &                          & Loss1 in Fig.~\ref{fig:fig1}       & \Checkmark        & \XSolidBrush                   &                             & 74.08 & -1.05        \\ \cline{3-5} \cline{7-8} 
                        &                          & Loss2 in Fig.~\ref{fig:fig1}        & \XSolidBrush         & \Checkmark                  &                             & \textbf{77.40} & \textbf{+2.27}        \\ \hline
\end{tabular}
\end{table*}
\subsection{Results on PASCAL VOC}
\label{S:4:s:4}
In this part, we train MobileNet-based YOLOv2 detectors on a more general and complex dataset named PASCAL VOC which includes 20 classes of objects to validate the robustness of our LRM method. Experiment results demonstrate that LRM method improves detection accuracy of YOLOv2 detector significantly. Table.~\ref{tab:tab3} shows all the details of this experiment and illustrates that all kinds of LRM-optimized YOLOv2 detectors outperform the original YOLOv2 detector. Improvements on PASCAL VOC demonstrate that LRM could help to construct more accurate real-time detectors in a wide range of real scenarios successfully, such as pedestrian detection, traffic monitoring and so forth. 

Totally, we construct 9 LRM-optimized YOLOv2 detectors with different hyperparameters on PASCAL VOC. Additionally, a MobileNet-based YOLOv2 detector without any additional optimization is used as the baseline. When setting the number of hard examples to 256 and the NMS threshold to 0.7, LRM method achieves the highest mAP(70.15\%) on PASCAL VOC dataset. Compared with the baseline whose mAP is 68.00\%, LRM method gains a significant improvements of over 2\% mAP.

More specifically, improvements that LRM brought to different classes are diverse. For some classes like bike, bottle and bus, the improvements are significant, achieving over 4\% mAP. But the case is opposite in a minority of classes, among which the typical ones are class cat and dog. For class cat, after applying LRM, detection accuracy of YOLOv2 drops moderately. However, for class, dog, detection accuracy drops significantly, achieving around 3\% mAP.

We utilize Fig.~\ref{fig:fig6} to illustrate the impacts brought by different hyperparameters on dataset PASCAL VOC. When setting the NMS threshold to 0.5 or 1.0(equals to no nms in Fig.~\ref{fig:fig6}), choosing 128 predictions for backpropagation achieves the best performance. But the case is totally different when the NMS threshold is set to 0.7. In this case, choosing 128 predictions perform worst(69.51\% mAP), but choosing 256 predictions achieves the highest mAP at 70.15\%. This is mainly because compared with NMS 0.5 and 1.0, setting NMS threshold to 0.7 is a milder strategy which remains a part of redundant informations. In this case, predictions ranked $65^{th}$ to $128^{th}$ possess a majority of redundant informations which does harm to the model. But when enlarge the number of hard examples to 256, more predictions containing beneficial informations are chosen and detection accuracy of the model increases significantly.

\subsection{Comparisons between Loss Rank Mining and Focal Loss}
\label{S:4:s:5}
In this part, we use Focal Loss method and our Loss Rank Mining method to optimize two YOLOv2 detectors, respectively. For clearer comparisons, darknet\cite{redmon2016you} rather than MobileNet\cite{howard2017mobilenets} is chosen as the backbone network. To make it possible for Focal Loss(FL) method to be applied to YOLOv2 detector, the original loss function defined as Loss2 in Fig.~\ref{fig:fig1} is replaced by Loss1 in Fig.~\ref{fig:fig1} in this part of experiments. 

Results are illustrated in Table.~\ref{tab:tab5}. Some details like hyperparamaters are not shown in this table. When Loss2 is replaced with Loss1, mAP drops with 2.08\%. Though after using FL method on this model, mAP increases with 1.03\%, mAP still drops with 1.05\% compared with the original YOLOv2 detector using Loss2(in Fig.~\ref{fig:fig1}) as its loss function. Compared with FL method, our Loss Rank Mining method could be applied to the original detector straightly and improve the detection accuracy significantly. Compared with the original YOLOv2 detector, the improvements are 2.27\%(77.40\% vs. 75.13\% mAP).

Though using Focal Loss method on YOLOv2 detector is possible, the prerequisite of changing the original loss function to another one does harm to detection accuracy. Even after applying Focal Loss to YOLOv2, the performance is still worse than the original one(74.05\% vs. 75.13\% mAP). Differently, our Loss Rank Mining can be applied to YOLOv2 detector straightly without any modification in loss function. More importantly, after using LRM method, YOLOv2 detector gains over 2\% mAP improvements on PASCAL VOC, compared the original one(77.40\% vs. 75.13\% mAP).

The reason why Loss Rank Mining method could mine foregrounds from tons of backgrounds can be explained as follows. When training real-time object detectors, the model incline to be partial to backgrounds and perform well on them. The corresponding result is that few foregrounds can be detected correctly, which is opposite the target of constructing object detectors. LRM method utilizes the fact that foregrounds tend to become hard examples. It ignores some well-predicted bounding-boxes in each iteration and make foregrounds dominate the process of gradient descent. Accordingly, LRM method emphasizes foregrounds successfully and helps to correct the deviation caused by huge imbalance between backgrounds and foregrounds. 

\section{Conclusion}
\label{S:5}

In this work, to begin with, we point out the strong demand and great importance of utilizing hard example mining strategies in real-time detectors. Then, to tackle the shortage of general hard example mining methods for real-time detectors, we propose the Loss Rank Mining method. Our method is a general one which does not rely on additional region proposals or specific loss functions, and it can be straightly used in state-of-the-art real-time detectors to boost their detection accuracy significantly. For demonstrating the effectiveness of LRM method, series of solid experiments are set. Results illustrate that no matter on auto-driving related dataset KITTI or more general dataset PASCAL VOC, LRM method boosts accuracy of real-time detectors significantly. In addition, broadly-used YOLOv2 detector which has yet to be optimized by previous hard example mining method could benefit from our method. However, we have noticed two shortcomings of our LRM method. One is that the fixed number of hard examples is unsuitable, and our future work will concentrate on making this hyperparameter self-adaptive. The other one is that filtering easy examples straightly may lose some valuable information. We will try to alleviate the impacts of easy examples instead of neglecting them in our future work.



%% References


\bibliographystyle{IEEEtran}
\bibliography{sample}

\begin{thebibliography}{10}
\providecommand{\url}[1]{#1}
\csname url@samestyle\endcsname
\providecommand{\newblock}{\relax}
\providecommand{\bibinfo}[2]{#2}
\providecommand{\BIBentrySTDinterwordspacing}{\spaceskip=0pt\relax}
\providecommand{\BIBentryALTinterwordstretchfactor}{4}
\providecommand{\BIBentryALTinterwordspacing}{\spaceskip=\fontdimen2\font plus
\BIBentryALTinterwordstretchfactor\fontdimen3\font minus
  \fontdimen4\font\relax}
\providecommand{\BIBforeignlanguage}[2]{{%
\expandafter\ifx\csname l@#1\endcsname\relax
\typeout{** WARNING: IEEEtran.bst: No hyphenation pattern has been}%
\typeout{** loaded for the language `#1'. Using the pattern for}%
\typeout{** the default language instead.}%
\else
\language=\csname l@#1\endcsname
\fi
#2}}
\providecommand{\BIBdecl}{\relax}
\BIBdecl

\bibitem{girshick2014rich}
R.~Girshick, J.~Donahue, T.~Darrell, and J.~Malik, ``Rich feature hierarchies
  for accurate object detection and semantic segmentation,'' in
  \emph{Proceedings of the IEEE conference on computer vision and pattern
  recognition}, 2014, pp. 580--587.

\bibitem{girshick2015fast}
R.~Girshick, ``Fast r-cnn,'' in \emph{Proceedings of the IEEE international
  conference on computer vision}, 2015, pp. 1440--1448.

\bibitem{ren2015faster}
S.~Ren, K.~He, R.~Girshick, and J.~Sun, ``Faster r-cnn: Towards real-time
  object detection with region proposal networks,'' in \emph{Advances in neural
  information processing systems}, 2015, pp. 91--99.

\bibitem{dai2016r}
J.~Dai, Y.~Li, K.~He, and J.~Sun, ``R-fcn: Object detection via region-based
  fully convolutional networks,'' in \emph{Advances in neural information
  processing systems}, 2016, pp. 379--387.

\bibitem{he2017mask}
K.~He, G.~Gkioxari, P.~Doll{\'a}r, and R.~Girshick, ``Mask r-cnn,'' \emph{arXiv
  preprint arXiv:1703.06870}, 2017.

\bibitem{liu2016ssd}
W.~Liu, D.~Anguelov, D.~Erhan, C.~Szegedy, S.~Reed, C.-Y. Fu, and A.~C. Berg,
  ``Ssd: Single shot multibox detector,'' in \emph{European conference on
  computer vision}.\hskip 1em plus 0.5em minus 0.4em\relax Springer, 2016, pp.
  21--37.

\bibitem{lin2017focal}
T.-Y. Lin, P.~Goyal, R.~Girshick, K.~He, and P.~Doll{\'a}r, ``Focal loss for
  dense object detection,'' \emph{arXiv preprint arXiv:1708.02002}, 2017.

\bibitem{redmon2016yolo9000}
J.~Redmon and A.~Farhadi, ``Yolo9000: better, faster, stronger,'' \emph{arXiv
  preprint arXiv:1612.08242}, 2016.

\bibitem{redmon2016you}
J.~Redmon, S.~Divvala, R.~Girshick, and A.~Farhadi, ``You only look once:
  Unified, real-time object detection,'' in \emph{Proceedings of the IEEE
  Conference on Computer Vision and Pattern Recognition}, 2016, pp. 779--788.

\bibitem{suykens1999least}
J.~A. Suykens and J.~Vandewalle, ``Least squares support vector machine
  classifiers,'' \emph{Neural processing letters}, vol.~9, no.~3, pp. 293--300,
  1999.

\bibitem{dollar2009integral}
P.~Doll{\'a}r, Z.~Tu, P.~Perona, and S.~Belongie, ``Integral channel
  features,'' 2009.

\bibitem{dalal2005histograms}
N.~Dalal and B.~Triggs, ``Histograms of oriented gradients for human
  detection,'' in \emph{Computer Vision and Pattern Recognition, 2005. CVPR
  2005. IEEE Computer Society Conference on}, vol.~1.\hskip 1em plus 0.5em
  minus 0.4em\relax IEEE, 2005, pp. 886--893.

\bibitem{he2014spatial}
K.~He, X.~Zhang, S.~Ren, and J.~Sun, ``Spatial pyramid pooling in deep
  convolutional networks for visual recognition,'' in \emph{European Conference
  on Computer Vision}.\hskip 1em plus 0.5em minus 0.4em\relax Springer, 2014,
  pp. 346--361.

\bibitem{shrivastava2016training}
A.~Shrivastava, A.~Gupta, and R.~Girshick, ``Training region-based object
  detectors with online hard example mining,'' in \emph{Proceedings of the IEEE
  Conference on Computer Vision and Pattern Recognition}, 2016, pp. 761--769.

\bibitem{lin2016feature}
T.-Y. Lin, P.~Doll{\'a}r, R.~Girshick, K.~He, B.~Hariharan, and S.~Belongie,
  ``Feature pyramid networks for object detection,'' \emph{arXiv preprint
  arXiv:1612.03144}, 2016.

\bibitem{redmon2013darknet}
J.~Redmon, ``Darknet: Open source neural networks in c,'' \emph{h
  ttp://pjreddie. com/darknet}, vol. 2016, 2013.

\bibitem{geiger2013vision}
A.~Geiger, P.~Lenz, C.~Stiller, and R.~Urtasun, ``Vision meets robotics: The
  kitti dataset,'' \emph{The International Journal of Robotics Research},
  vol.~32, no.~11, pp. 1231--1237, 2013.

\bibitem{everingham2010pascal}
M.~Everingham, L.~Van~Gool, C.~K. Williams, J.~Winn, and A.~Zisserman, ``The
  pascal visual object classes (voc) challenge,'' \emph{International journal
  of computer vision}, vol.~88, no.~2, pp. 303--338, 2010.

\bibitem{deng2009imagenet}
J.~Deng, W.~Dong, R.~Socher, L.-J. Li, K.~Li, and L.~Fei-Fei, ``Imagenet: A
  large-scale hierarchical image database,'' in \emph{Computer Vision and
  Pattern Recognition, 2009. CVPR 2009. IEEE Conference on}.\hskip 1em plus
  0.5em minus 0.4em\relax IEEE, 2009, pp. 248--255.

\bibitem{wang2016factorized}
M.~Wang, B.~Liu, and H.~Foroosh, ``Factorized convolutional neural networks,''
  \emph{arXiv preprint arXiv:1608.04337}, 2016.

\bibitem{iandola2016squeezenet}
F.~N. Iandola, S.~Han, M.~W. Moskewicz, K.~Ashraf, W.~J. Dally, and K.~Keutzer,
  ``Squeezenet: Alexnet-level accuracy with 50x fewer parameters and< 0.5 mb
  model size,'' \emph{arXiv preprint arXiv:1602.07360}, 2016.

\bibitem{wu2016quantized}
J.~Wu, C.~Leng, Y.~Wang, Q.~Hu, and J.~Cheng, ``Quantized convolutional neural
  networks for mobile devices,'' in \emph{Proceedings of the IEEE Conference on
  Computer Vision and Pattern Recognition}, 2016, pp. 4820--4828.

\bibitem{howard2017mobilenets}
A.~G. Howard, M.~Zhu, B.~Chen, D.~Kalenichenko, W.~Wang, T.~Weyand,
  M.~Andreetto, and H.~Adam, ``Mobilenets: Efficient convolutional neural
  networks for mobile vision applications,'' \emph{arXiv preprint
  arXiv:1704.04861}, 2017.

\bibitem{zhang2017shufflenet}
X.~Zhang, X.~Zhou, M.~Lin, and J.~Sun, ``Shufflenet: An extremely efficient
  convolutional neural network for mobile devices,'' \emph{arXiv preprint
  arXiv:1707.01083}, 2017.

\end{thebibliography}

\end{document}